\pdfoutput=1

\documentclass[11pt]{article}

\usepackage[final]{acl}

\usepackage{times}
\usepackage{latexsym}
\usepackage{amsmath}

\usepackage[T1]{fontenc}

\usepackage[utf8]{inputenc}

\usepackage{microtype}

\usepackage{inconsolata}

\usepackage{graphicx}

\title{FlavorDiffusion: Predicting Food Pairings and Chemical Interactions Using Diffusion Models}

\author{Jun Pyo, Seo \\
  Seoul National University \\ 
  jpseo99\texttt{@snu.ac.kr}}

\begin{document}
\maketitle

\begin{abstract}
The study of food pairing has evolved beyond subjective expertise with the advent of machine learning. This paper presents FlavorDiffusion, a novel framework leveraging diffusion models to predict food-chemical interactions and ingredient pairings without relying on chromatography. By integrating graph-based embeddings \cite{Perozzi2014}, diffusion processes \cite{Ho2020, Song2021, Sun2023}, and chemical property encoding \cite{Azambuja2023}, FlavorDiffusion addresses data imbalances and enhances clustering quality. Using a heterogeneous graph derived from datasets like Recipe1M \cite{Marin2019} and FlavorDB, our model demonstrates superior performance in reconstructing ingredient-ingredient relationships. The addition of a Chemical Structure Prediction (CSP) layer further refines the embedding space, achieving state-of-the-art NMI scores and enabling meaningful discovery of novel ingredient combinations. The proposed framework represents a significant step forward in computational gastronomy, offering scalable, interpretable, and chemically informed solutions for food science.
\end{abstract}

\section{Introduction}
Food pairing has traditionally relied on the intuition and experience of chefs, yet scientific analysis and optimization of food combinations remain underexplored. Recent research has leveraged data-driven approaches to model the relationships between food ingredients and chemical compounds to predict novel food pairings.

Several computational approaches have been developed to model food pairings and ingredient relationships. {Kitchenette} \cite{Park2019}, for instance, applies Siamese neural networks to predict and recommend ingredient pairings based on a large annotated dataset. However, it suffers from key limitations, such as a lack of chemical interpretability and heavy reliance on labeled data, making it less generalizable across different cuisines and novel food combinations.

One of the key advancements in this domain is {FlavorGraph} \cite{Park2019}, a large-scale food-chemical deep neural network model comprising {6,653 ingredient nodes} and {1,645 compound nodes}. This graph captures two primary relationships: (1) {ingredient-ingredient relations}, representing co-occurrence patterns in recipes, and (2) {ingredient-compound relations}, indicating chemical composition links. These relationships are constructed using datasets such as {Recipe1M} \cite{Marin2019}, {FlavorDB}, and {HyperFoods}. {FlavorGraph} incorporates food-chemical associations into a neural network by leveraging the {metapath2vec} \cite{Dong2017} algorithm, which embeds ingredient-compound relationships in a word2vec-like manner. Expanding on this approach, {WineGraph} \cite{Gawrysiak2023} extends the framework by integrating wine-related datasets to define optimal food-wine pairings.

Despite progress in computational food science, major challenges remain. Chromatography-based methods, while precise, are costly and limit the acquisition of large-scale chemical interaction data. {FlavorGraph} effectively captures ingredient-compound relationships using {metapath-based embeddings}, but its reliance on {random-walk sampling} makes it difficult to incorporate edge weights and spatial information within the graph structure. These limitations hinder the full exploitation of food-chemical associations, leading to suboptimal ingredient relationship modeling. To address these challenges, we introduce {FlavorDiffusion}, a Diffusion Model-based framework that refines the representation of food-chemical interactions and elevates the quality of food pairing predictions.

\subsection*{Contributions}
\begin{itemize}
    \item We propose a {graph-based diffusion modeling approach} that leverages {DIFUSCO} \cite{Sun2023} to capture richer and more structured representations of food-chemical interactions.
    \item We introduce a {balanced subgraph sampling} strategy to address data imbalance issues, ensuring fair representation across different ingredient-chemical associations.
    \item Our experimental results demonstrate improvements in {Normalized Pointwise Mutual Information (NPMI)} scores for node embeddings, facilitating more effective chemical inference.
    \item We establish a foundation for predicting chromatography results for {non-hub chemicals}, extending the applicability of our model beyond frequently occurring compounds.
    \item Our approach enables {pairing inference using chemical properties}, providing structured and interpretable recommendations for novel ingredient combinations.
\end{itemize}

\section{Dataset}

Our study builds upon {FlavorGraph} \cite{Park2019} by utilizing the same large-scale datasets to construct a robust food-chemical network. These datasets provide a structured representation of ingredient relationships and chemical interactions. In the following sections, we summarize the key characteristics of these datasets and outline the preprocessing steps applied to ensure data consistency and usability in our framework.

\begin{table}[h]
  \centering
  \begin{tabular}{lccc}
    \hline
    {Type} & {Source} & {Nodes} & {Edges} \\
    \hline
    I-I & Recipe1M & 6,653 & 111,355 \\
    I-FC & FlavorDB & 1,561 & 35,440 \\
    I-DC & HyperFoods & 84 & 386 \\
    \hline
    {Total} & - & 8,298 & 147,179 \\
    \hline
  \end{tabular}
  \caption{Summary of the heterogeneous food-compound graph. I-I represents ingredient ingredient co-occurrence from Recipe1M, I-FC denotes ingredient-flavor compound associations from FlavorDB, and I-DC refers to ingredient-drug compound relations}

  \label{tab:graph_summary}
\end{table}

\subsection{Data Sources}
This study utilizes the same datasets as {FlavorGraph} \cite{Park2019} to construct a structured food-chemical network.

{Recipe1M} \cite{Marin2019} contains {65,284} recipes with ingredient lists and cooking instructions, capturing ingredient co-occurrence patterns in real-world culinary practices.

{FlavorDB} compiles chemical composition data from multiple sources, including \textit{FooDB}, \textit{Flavornet}, and \textit{BitterDB}. It originally includes {2,254 flavor compounds} linked to {936 food ingredients}, but only {400 commonly used ingredients} were selected to align with Recipe1M, resulting in {1,561 flavor compound nodes} and {164,531 ingredient-flavor compound edges}.

{HyperFoods} maps drug compounds to food ingredients using machine learning based on food-gene interactions. From the original {206 food ingredients}, {104 were selected}, yielding {84 drug compound nodes} and {386 ingredient-drug compound edges}.

\subsection{Data Processing}

To construct a structured representation of food-chemical relationships, we build upon {FlavorGraph} \cite{Park2019}, a heterogeneous graph that integrates both culinary and chemical associations. The graph construction process follows a structured approach. First, an ingredient-ingredient graph is built by extracting co-occurrence patterns from Recipe1M \cite{Marin2019}, where edges between ingredients are established based on their {Normalized Pointwise Mutual Information (NPMI)} scores. Only statistically significant ingredient pairs appearing together in a substantial number of recipes are retained, resulting in a total of {111,355 edges}. Second, an ingredient-chemical graph is formed by linking ingredients to their corresponding chemical compounds using FlavorDB and HyperFoods, leading to {35,440 edges} between food ingredients and known chemical compounds. The final graph structure comprises {6,653 ingredient nodes} and {1,645 compound nodes}, forming a {heterogeneous graph} that encodes both culinary co-occurrence relationships and chemical interactions.

\subsection{Chemical Property Encoding}
To ensure chemically informed ingredient representations, each compound is characterized using {CACTVS chemical fingerprints}, which are encoded as {881-dimensional binary vectors}. These vectors represent molecular descriptors such as molecular weight, functional groups, and substructure patterns, using a {binary encoding scheme} where each bit indicates the presence or absence of a specific chemical substructure.

\section{Related Work}

\subsection{FlavorGraph}

FlavorGraph \cite{Park2019} is a heterogeneous graph \( G = (V, E) \) integrating ingredient co-occurrence and molecular profiling to model food-chemical interactions. By leveraging metapath-based learning \cite{Dong2017}, it enables systematic ingredient discovery and predictive food pairing through shared molecular properties.

\subsubsection{Metapath2Vec}

To learn chemically meaningful embeddings, we employ \textbf{Metapath2Vec}, which captures high-order relations via structured random walks. Ingredients are classified into hub ingredients (\( H \)), which directly connect to chemical compounds, and non-hub ingredients (\( N \)), which lack direct chemical links and rely on hub ingredients to acquire chemical insights.

The metapath sampling strategy follows:

\begin{equation*}
    N \rightarrow H \rightarrow C \rightarrow H \rightarrow N
\end{equation*}

where \( C \) represents chemical compounds. This structured propagation ensures that non-hub ingredients inherit chemical relevance, enhancing embedding robustness and interpretability.

\subsubsection{Architecture}

The network, parameterized by \( \theta \), takes node pairs \( (i, j) \) as input and outputs an edge score \( s_{\theta}(i, j) \), normalized across all embeddings:

\begin{equation*}
    s_{\theta}(i, j) = \sigma(\mathbf{u}_i^T \mathbf{u}_j)
\end{equation*}

where \( \mathbf{u}_i \) and \( \mathbf{u}_j \) are the learned embeddings for nodes \( i \) and \( j \), ensuring consistency across culinary co-occurrence and chemical similarity.

\subsubsection{Loss Function}

Embeddings are optimized using Skip-Gram with Negative Sampling (SGNS):

\begin{equation*}
    J_{\theta} = \sum_{(i, j) \in D} \log \sigma(\mathbf{u}_i^T \mathbf{u}_j) + \sum_{(i, j') \in D'} \log \sigma(-\mathbf{u}_i^T \mathbf{u}_{j'})
\end{equation*}

where \( D \) and \( D' \) are positive and negative sample pairs. To enforce chemical relevance, an additional \textbf{Chemical Structure Prediction (CSP)} loss is introduced:

\begin{equation*}
    L_{\text{CSP}, \theta} = \sum_{d=1}^{D} y_d \log f_{\theta, d}(i) + (1 - y_d) \log (1 - f_{\theta, d}(i))
\end{equation*}

where \( f_{\theta, d}(i) \) predicts the presence of the \( d \)-th molecular substructure \( y_d \), refining embeddings with molecular fingerprints.

\subsection{DIFUSCO}

Graph-based diffusion models have recently emerged as powerful frameworks for solving combinatorial optimization problems by leveraging probabilistic generative processes. In our work, we leverage the fundamental principles of graph-based diffusion models, particularly the Gaussian diffusion framework, to reconstruct structured graph representations. By incorporating diffusion-driven embeddings into our heterogeneous network, we enhance the predictive accuracy of food-chemical interactions while maintaining interpretability. This approach allows for the seamless integration of molecular-level insights into ingredient pairing research, further advancing computational gastronomy.

\section{Proposition: FlavorDiffusion}

\subsection{Sub-Graph Sampling}

FlavorDiffusion is built upon the DIFUSCO Gaussian noise-based diffusion model, extending its capabilities to structured food-chemical graphs. The core objective is to train a model capable of reconstructing subgraphs sampled from the full heterogeneous graph \( G = (V, E) \) while leveraging node attributes as guidance. 

The full graph consists of a diverse set of nodes \( V \), including hub ingredients, non-hub ingredients, flavor compounds, and drug compounds, with edges \( E \) encoding the strength of their relationships as continuous values in \( [0,1] \). We define a dataset of subgraphs, where each sample contains \( m \) nodes selected from \( G \). These subgraphs are denoted as:

\begin{equation*}
    \mathcal{D}_m = \{ G_i = (V_i, E_i) \}_{i=1}^{N},
\end{equation*}

where each subgraph \( G_i \) has \( |V_i| = m \) nodes and an adjacency matrix \( E_i \) of size \( m \times m \), representing pairwise edge scores. The dataset is partitioned into training (\( N_t \)) and validation (\( N_v \)) subsets.

\subsection{Forward Diffusion Process}

For a single data point \( G_i = (V_i, E_i) \) sampled from the dataset, we define the diffusion process over its edge set \( E_i \). By convention, we denote the corrupted version of \( E_i \) at timestep \( t \) as \( x_t \), aligning with standard diffusion formalisms. The node representations, encompassing all vertex features, are denoted as \( \mathbf{Emb} \).

The forward diffusion process follows a Markovian Gaussian noise injection, progressively perturbing the edges \( x_t \) while preserving node representations:

\begin{equation*}
    q(x_t | x_{t-1}) = \mathcal{N}(x_t; \sqrt{1 - \beta_t} x_{t-1}, \beta_t I),
\end{equation*}

where \( \beta_t \) is a predefined noise variance at timestep \( t \). Given an initial clean edge matrix \( x_0 = E_i \), we can analytically express the direct corruption of \( x_0 \) at any timestep \( t \) as:

\begin{equation*}
    q(x_t | x_0) = \mathcal{N}(x_t; \sqrt{\bar{\alpha}_t} x_0, (1 - \bar{\alpha}_t) I),
\end{equation*}

where \( \bar{\alpha}_t = \prod_{s=1}^{t} (1 - \beta_s) \) represents the cumulative noise effect over time. This formulation allows direct sampling of \( x_t \) from \( x_0 \), bypassing iterative updates.

In this framework, the edge structure is progressively degraded into Gaussian noise, while node representations \( \mathbf{Emb} \) remain unchanged, ensuring that denoising relies on learned node attributes.

\subsection{Reverse Denoising Process}

The reverse process seeks to recover \( x_0 \) from the fully corrupted state \( x_T \), learning to remove noise in a stepwise manner. The key assumption is that the forward process follows a Gaussian transition, enabling an analytically derived reverse process.

Given the Markovian nature of the diffusion process, we define the true posterior:

\begin{equation*}
    q(x_{t-1} | x_t, x_0) = \mathcal{N}(x_{t-1}; \tilde{\mu}_t(x_t, x_0), \tilde{\beta}_t I),
\end{equation*}

where the posterior mean and variance are derived as:

\begin{equation*}
    \tilde{\mu}_t(x_t, x_0) = \frac{\sqrt{\bar{\alpha}_{t-1}} \beta_t}{1 - \bar{\alpha}_t} x_0 + \frac{\sqrt{\alpha_t} (1 - \bar{\alpha}_{t-1})}{1 - \bar{\alpha}_t} x_t,
\end{equation*}

\begin{equation*}
    \tilde{\beta}_t = \frac{1 - \bar{\alpha}_{t-1}}{1 - \bar{\alpha}_t} \beta_t.
\end{equation*}

Since \( x_0 \) is unknown, we train a model \( p_{\theta}(x_0 | x_t) \) to approximate it. Substituting the predicted \( x_0 \), the learned reverse process is modeled as:

\begin{align*}
    p_{\theta}(x_{t-1} | x_t, \mathbf{Emb}) &= \\
    \mathcal{N}\big(x_{t-1}; &\mu_{\theta}(x_t, t, \mathbf{Emb}), \Sigma_{\theta}(x_t, t)\big),
\end{align*}

where \( \mu_{\theta} \) is the learned estimate for \( \tilde{\mu}_t(x_t, x_0) \), and the variance term is fixed as \( \Sigma_{\theta}(x_t, t) = \tilde{\beta}_t I \), avoiding the need for explicit learning. The function \( \mu_{\theta} \) is now conditioned on the node representations (\(\mathbf{Emb}\)) of the two vertices forming the edge.

Using the DDPM convention, we parameterize \( \mu_{\theta} \) as:

\begin{align*}
    \mu_{\theta}(x_t, t, \mathbf{Emb}) &= \frac{1}{\sqrt{\alpha_t}} \Bigg( x_t \\
    &\quad - \frac{\beta_t}{\sqrt{1 - \bar{\alpha}_t}} \epsilon_{\theta}(x_t, t, \mathbf{Emb}) \Bigg),
\end{align*}

where \( \epsilon_{\theta}(x_t, t, \mathbf{Emb}) \) is the learned noise estimate, which is now explicitly conditioned on the representations of the two nodes forming the edge. The node representations provide additional context for denoising by leveraging node-specific features.

\subsection{Optimization via Variational Lower Bound}

To train the reverse model, we maximize the variational lower bound (ELBO), decomposed as:

\begin{align*}
    \mathcal{L}_{\text{ELBO}} = E_q \Bigg[
    \log p_{\theta}(x_0 | x_1, \mathbf{Emb}) \\
    - \sum_{t=1}^{T} D_{\text{KL}}\big(q(x_{t-1} | x_t, x_0) \| p_{\theta}(x_{t-1} | x_t, \mathbf{Emb})\big) 
    \Bigg].
\end{align*}

Here, \( T \) represents the total number of diffusion steps, defining the depth of the forward and reverse process. The KL divergence encourages the learned transitions to match the true posterior. Since \( q(x_t | x_0) \) is Gaussian, minimizing \( D_{\text{KL}} \) is equivalent to predicting the noise component \( \epsilon \) added during diffusion. Thus, the training objective simplifies to:

\begin{equation*}
    \mathcal{L}_{\text{recon}} = E_{t, x_0, \epsilon} \left[ \| \epsilon - \epsilon_{\theta}(x_t, t, \mathbf{Emb}) \|^2 \right].
\end{equation*}

This loss ensures that \( \epsilon_{\theta} \) effectively estimates the noise introduced in the forward process while incorporating node representations. By iteratively refining the denoising function, FlavorDiffusion reconstructs the original ingredient-ingredient graph from noisy subgraphs, leveraging both the structural edge information and node attributes to enhance predictive modeling for food pairing analysis.

\subsection{Inference}

Graph reconstruction follows Denoising Diffusion Implicit Models (DDIM) for efficient and deterministic sampling. Unlike DDPM, DDIM removes noise via a non-Markovian update, accelerating inference.

Starting from \( x_T \sim \mathcal{N}(0, I) \), the reverse process iterates:

\begin{equation*}
    x_{t-1} = \sqrt{\bar{\alpha}_{t-1}} \hat{x}_0 + \sqrt{1 - \bar{\alpha}_{t-1}} \cdot \epsilon_{\theta}(x_t, t, \mathbf{Emb}),
\end{equation*}

where the predicted clean graph is:

\begin{equation*}
    \hat{x}_0 = \frac{x_t - \sqrt{1 - \bar{\alpha}_t} \epsilon_{\theta}(x_t, t, \mathbf{Emb})}{\sqrt{\bar{\alpha}_t}}.
\end{equation*}

Iterating from \( T \) to \( 0 \), the model refines \( x_t \) to recover ingredient-ingredient relationships. DDIM ensures fast, stable, and chemically meaningful reconstructions.

\subsection{Model Architecture}

The noise prediction network \( \epsilon_{\theta}(x_t, t, \mathbf{V}) \) employs an anisotropic GNN to iteratively refine node and edge embeddings. Let \( h_i^\ell \in \mathbf{R}^d \) and \( e_{ij}^\ell \in \mathbf{R}^{d_e} \) denote the node and edge features at layer \( \ell \), respectively. The refinement process updates both edge and node embeddings through the following operations:

\paragraph{Edge Refinement} The initial edge embeddings \( e_{ij}^0 \) are set as the corresponding values from the noisy edge representation \( x_t \). At each layer \( \ell \), the intermediate edge embeddings \( \hat{e}_{ij}^\ell \) are updated as:

\begin{equation*}
    \hat{e}_{ij}^\ell = P^\ell e_{ij}^\ell + Q^\ell h_i^\ell + R^\ell h_j^\ell,
\end{equation*}

where \( P^\ell, Q^\ell, R^\ell \in \mathbf{R}^{d_e \times d_e} \) are learnable parameters. The refined edge embedding \( e_{ij}^{\ell+1} \) is then computed as:

\begin{equation*}
    e_{ij}^{\ell+1} = e_{ij}^\ell + \text{MLP}_e\big(\text{BN}(\hat{e}_{ij}^\ell)\big) + \text{MLP}_t(t),
\end{equation*}

where \( \text{MLP}_e \) is a 2-layer perceptron and \( \text{MLP}_t \) embeds the diffusion timestep \( t \) using sinusoidal features.

\paragraph{Node Refinement}
The node embeddings \( h_i^\ell \) are refined by aggregating information from neighboring nodes and their associated edges. The update rule for \( h_i^{\ell+1} \) is given by:

\begin{equation*}
    h_i^{\ell+1} = h_i^\ell + \alpha \cdot \text{BN}\Big(U^\ell h_i^\ell + \sum_{j \in \mathcal{N}(i)} \sigma(\hat{e}_{ij}^\ell) \odot V^\ell h_j^\ell\Big),
\end{equation*}

where \( U^\ell, V^\ell \in \mathbf{R}^{d \times d} \) are learnable parameter matrices, \( \sigma \) is the sigmoid activation function used for edge gating, \( \odot \) denotes the Hadamard (element-wise) product, \( \mathcal{N}(i) \) represents the set of neighbors for node \( i \), and \( \alpha \) is the ReLU activation applied after aggregation.

\paragraph{Final Prediction}
After \( L \) GNN layers, the final refined edge embeddings \( E^{(L)} \in \mathbf{R}^{N \times N \times d_e} \) are passed through a ReLU activation and a multi-layer perceptron (MLP) to predict the noise:

\begin{equation*}
    \epsilon_{\theta}(x_t, t, \mathbf{V}) = \text{MLP}\big(\text{ReLU}(E^{(L)})\big).
\end{equation*}

This formulation ensures that both node and edge embeddings are iteratively refined to capture local and global graph structure, enabling robust denoising and reconstruction of ingredient-ingredient relationships.

\section{Experimental Results}

The evaluation consists of two primary experiments: (1) reproducing the NMI-based clustering performance evaluation originally conducted in FlavorGraph, and (2) assessing the generalization ability of our proposed Flavor Diffusion framework by testing on subgraphs of different sizes.

Subgraphs of size 25, 50, 100, and 200 nodes were sampled while maintaining an equal proportion of hub and non-hub ingredients. The number of subgraphs used for training and testing at each scale is shown in Table~\ref{tab:subgraph_stats}. 

\begin{table}[h!]
  \centering
  \caption{Subgraph Composition for Training and Testing}
  \label{tab:subgraph_stats}
  \resizebox{\linewidth}{!}{%
  \begin{tabular}{ccc}
    \hline
    \textbf{Nodes per Subgraph} & \textbf{Train Set Size} & \textbf{Test Set Size} \\
    \hline
    25  & 256,000 & 256 \\
    50  & 128,000 & 128 \\
    100 & 64,000  & 64  \\
    200 & 32,000  & 32  \\
    \hline
  \end{tabular}%
  }
\end{table}

\paragraph{Generalization Ability}

To assess the generalization ability of the proposed framework, models trained on one subgraph size were tested on all sizes to observe performance across different scales. The results in Table~\ref{tab:generalization_results} indicate that models trained on 25-node subgraphs generalize poorly to larger graphs, with an MSE of 0.025078 when tested on 100-node subgraphs. In contrast, the 100-node trained model demonstrates the most stable generalization across different test sizes, showing minimal MSE variation. The 200-node trained model, while excelling on large graphs with an MSE of 0.003692, exhibits difficulties in adapting to smaller structures, with a high error of 0.059557 when tested on 25-node subgraphs.

\begin{table}[h!]
  \centering
  \caption{Generalization Performance: Validation MSE Loss}
  \label{tab:generalization_results}
  \resizebox{\linewidth}{!}{%
  \begin{tabular}{lcccc}
    \hline
    \textbf{Train Size} & \textbf{Test (25)} & \textbf{Test (50)} & \textbf{Test (100)} & \textbf{Test (200)} \\
    \hline
    25  & 0.004589 & 0.010965 & 0.025078 & 0.019477 \\
    50  & 0.025235 & 0.005884 & 0.004420 & 0.004123 \\
    100 & 0.003964 & 0.003678 & 0.004232 & 0.003953 \\
    200 & \textbf{0.059557} & 0.007837 & 0.003992 & \textbf{0.003692} \\
    \hline
  \end{tabular}%
  }
\end{table}

These results highlight that subgraph size significantly impacts both intra-subgraph clustering and cross-subgraph generalization performance. The Flavor Diffusion (100 nodes) model provides the best balance between clustering accuracy and scalability, demonstrating the ability to generalize well across varying ingredient graph structures. On the other hand, training on extremely small subgraphs limits generalization, while models trained on large subgraphs struggle when applied to smaller ingredient sets. These findings suggest that a mid-sized subgraph training approach (e.g., 100 nodes) is optimal for robust ingredient representation learning.

\paragraph{NMI-based Evaluation}

To construct the clustering test dataset, nine representative food categories were defined: \textit{Bakery/Dessert/Snack, Beverage Alcoholic, Cereal/Crop/Bean, Dairy, Fruit, Meat/Animal Product, Plant/Vegetable, Seafood, and Others}. From these, 416 chemical hub ingredients with strong connections were selected to ensure diverse and well-defined clustering labels, enabling fair comparisons across models commonly used in related studies.

The NMI-based evaluation results in Table~\ref{tab:nmi_results} demonstrate the clustering quality of different models. Among the non-CSP variants, the Flavor Diffusion (50 nodes) model achieves the highest NMI score of 0.3236, surpassing the baseline FlavorGraph model without CSP. The best overall performance is observed in the Flavor Diffusion\_CSP (200 nodes) model, which achieves an NMI score of 0.3410, indicating that the CSP layer significantly improves the learned ingredient embeddings. Smaller subgraphs, such as the 25-node configuration, show the greatest improvement when using CSP (0.2970 vs. 0.2167), suggesting that the chemical structure prediction enhances clustering, particularly in more limited ingredient sets.

\begin{table}[h!]
  \centering
  \caption{Performance Comparison Using NMI Metric. *CSP shorts for chemical structure prediction.}
  \label{tab:nmi_results}
  \resizebox{\linewidth}{!}{%
  \begin{tabular}{lcc}
    \hline
    \textbf{Model} & \textbf{NMI Mean} & \textbf{NMI Std} \\
    \hline
    FlavorGraph \cite{Park2019}      & $0.2995$ & $0.0403$ \\
    FlavorGraph\_CSP \cite{Park2019} & $0.3102$ & $0.0407$ \\
    \hline
    Flavor Diffusion (25 nodes)       & $0.2167$ & $0.0319$ \\
    Flavor Diffusion (50 nodes)       & $\mathbf{0.3236}$ & $\mathbf{0.0134}$ \\
    Flavor Diffusion (100 nodes)      & $0.3170$ & $0.0207$ \\
    Flavor Diffusion (200 nodes)      & $0.2935$ & $0.0300$ \\
    \hline
    Flavor Diffusion\_CSP (25 nodes)  & $0.2970$ & $0.0144$ \\
    Flavor Diffusion\_CSP (50 nodes)  & $0.2862$ & $0.0152$ \\
    Flavor Diffusion\_CSP (100 nodes) & $0.3169$ & $0.0257$ \\
    Flavor Diffusion\_CSP (200 nodes) & $\mathbf{0.3410}$ & $\mathbf{0.0150}$ \\
    \hline
  \end{tabular}%
  }
\end{table}
\section{Discussion}

\begin{figure*}[t!]
  \centering
  \includegraphics[width=0.32\linewidth]{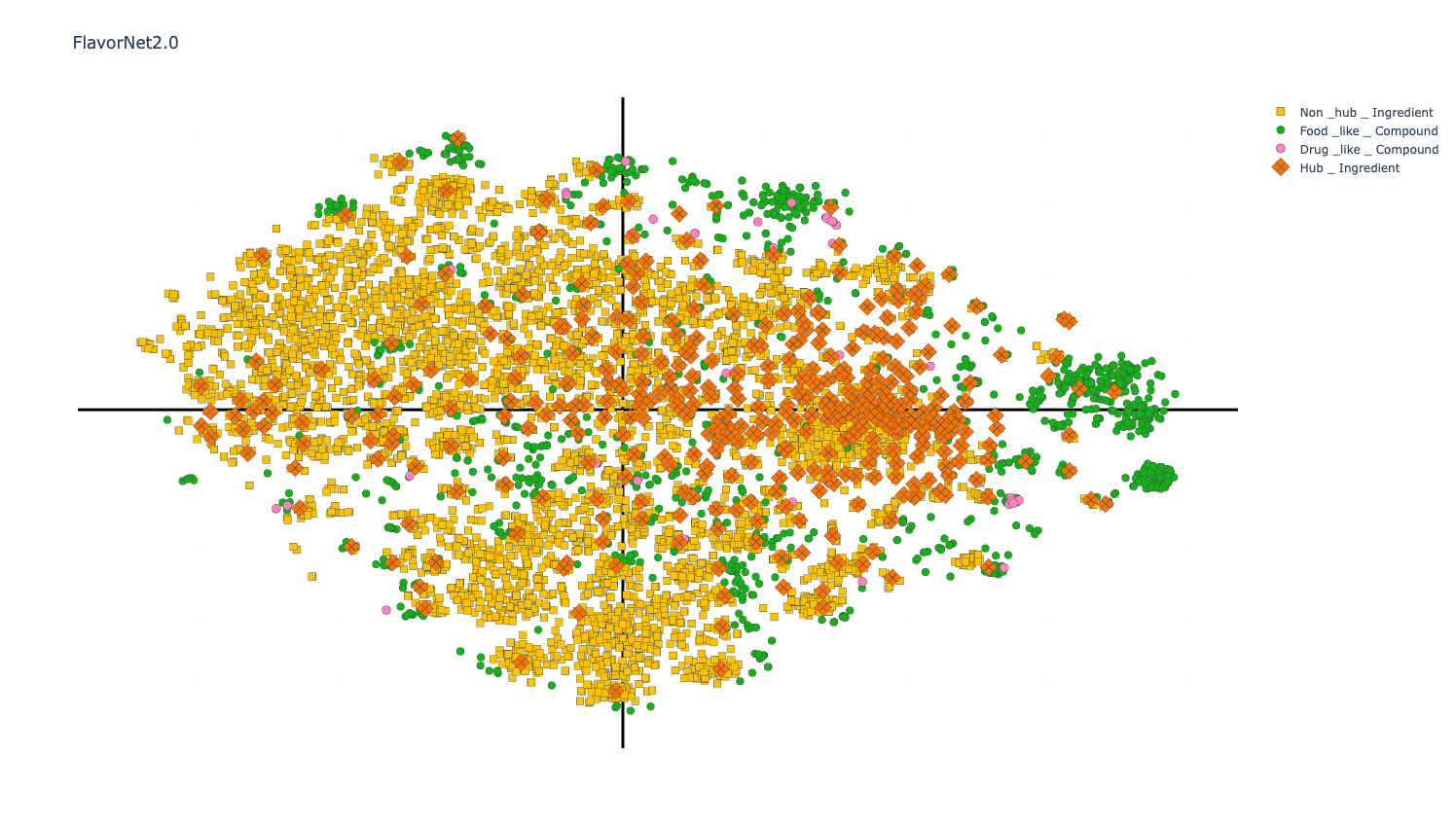} \hfill
  \includegraphics[width=0.32\linewidth]{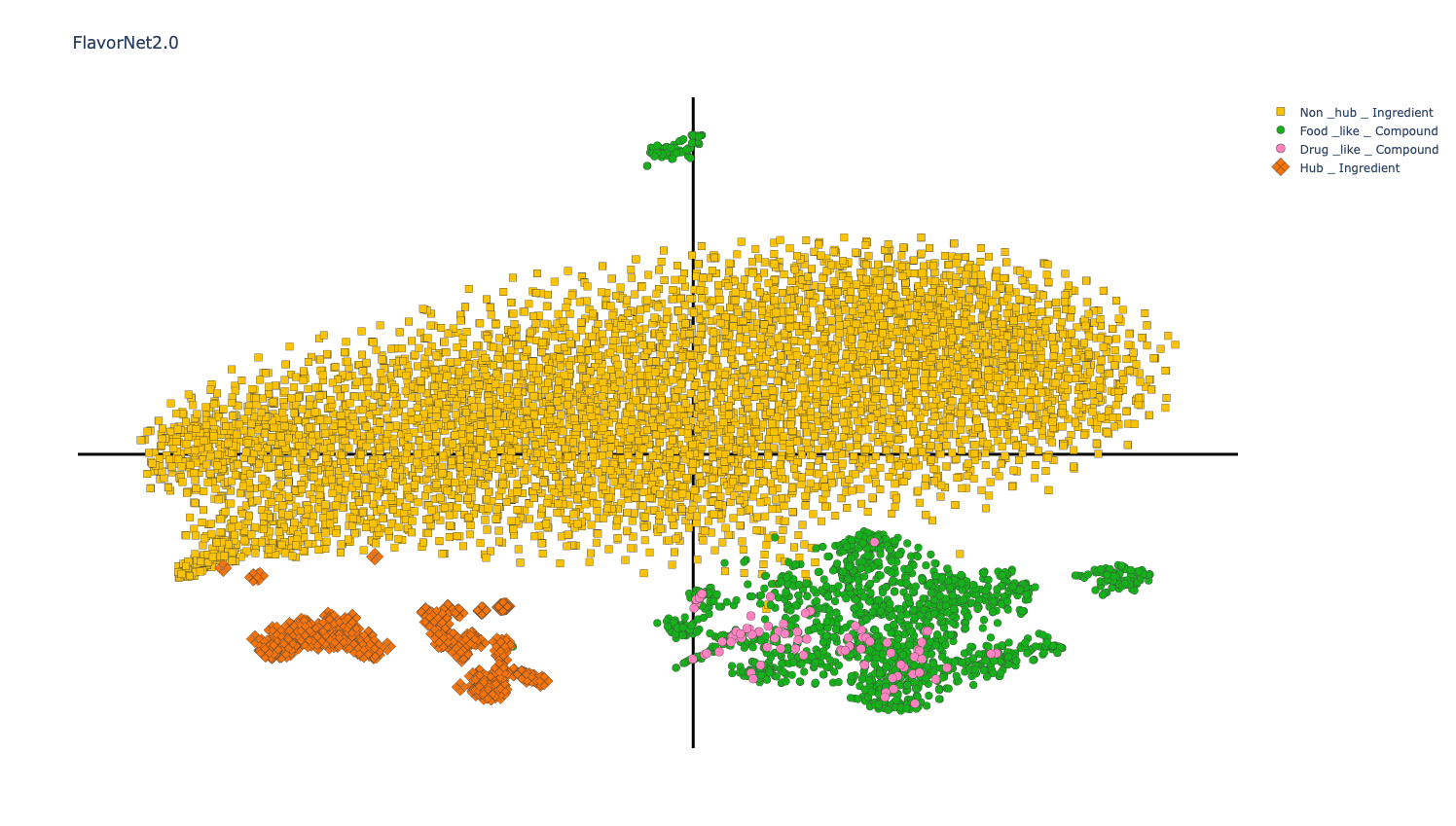} \hfill
  \includegraphics[width=0.32\linewidth]{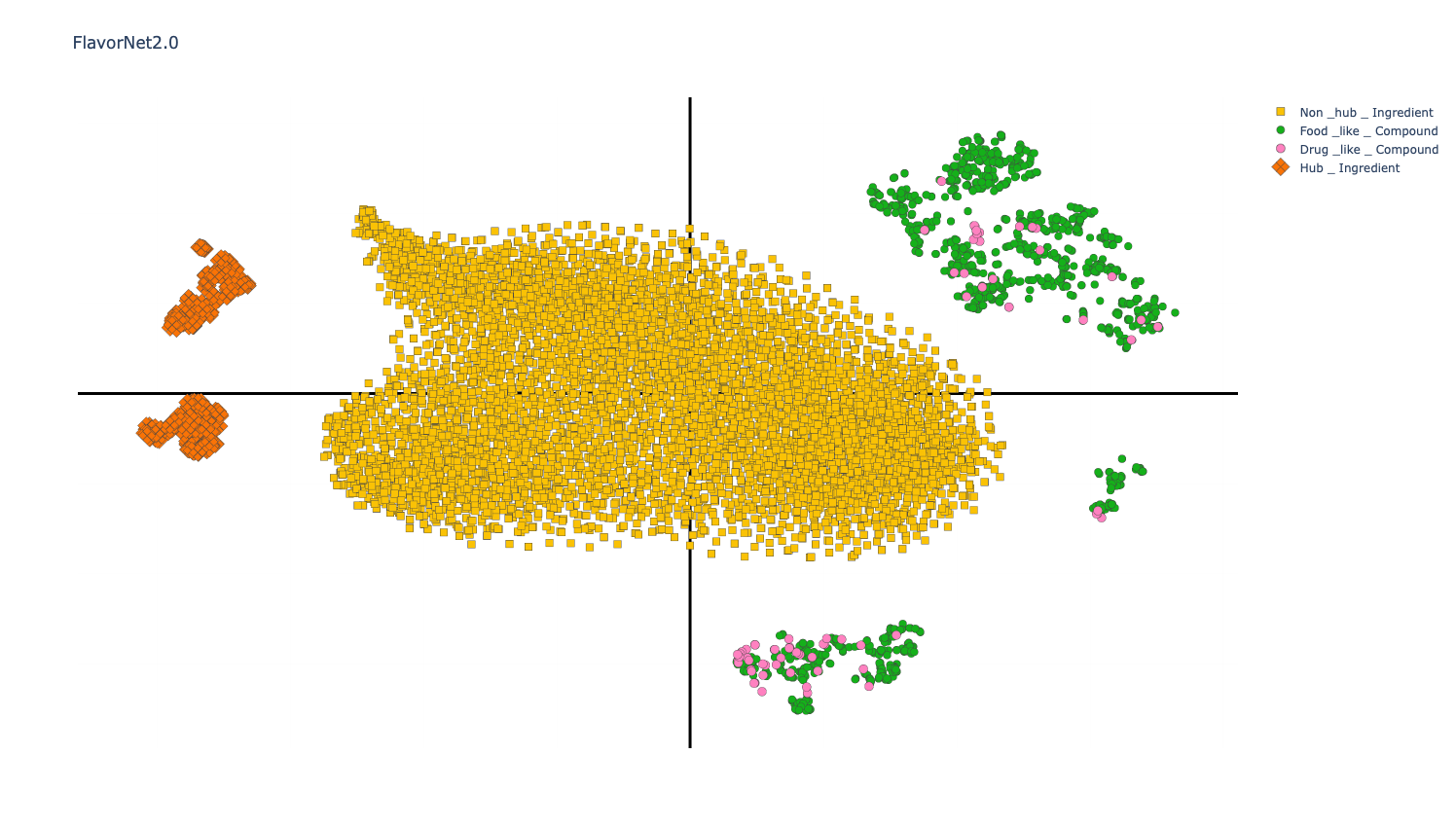}
  \caption{Embedding space comparison under different configurations. 
  (Left) Baseline embeddings show poor separation between ingredients and compounds. 
  (Center) Flavor Diffusion (200 nodes) without CSP achieves improved clustering of chemical compounds and hub ingredients. 
  (Right) Flavor Diffusion (200 nodes) with CSP results in well-defined clusters, leveraging chemical fingerprints to enhance separation.}
  \label{fig:embedding_comparison}
\end{figure*}

\begin{figure}[h!]
    \centering
    \includegraphics[width=\linewidth]{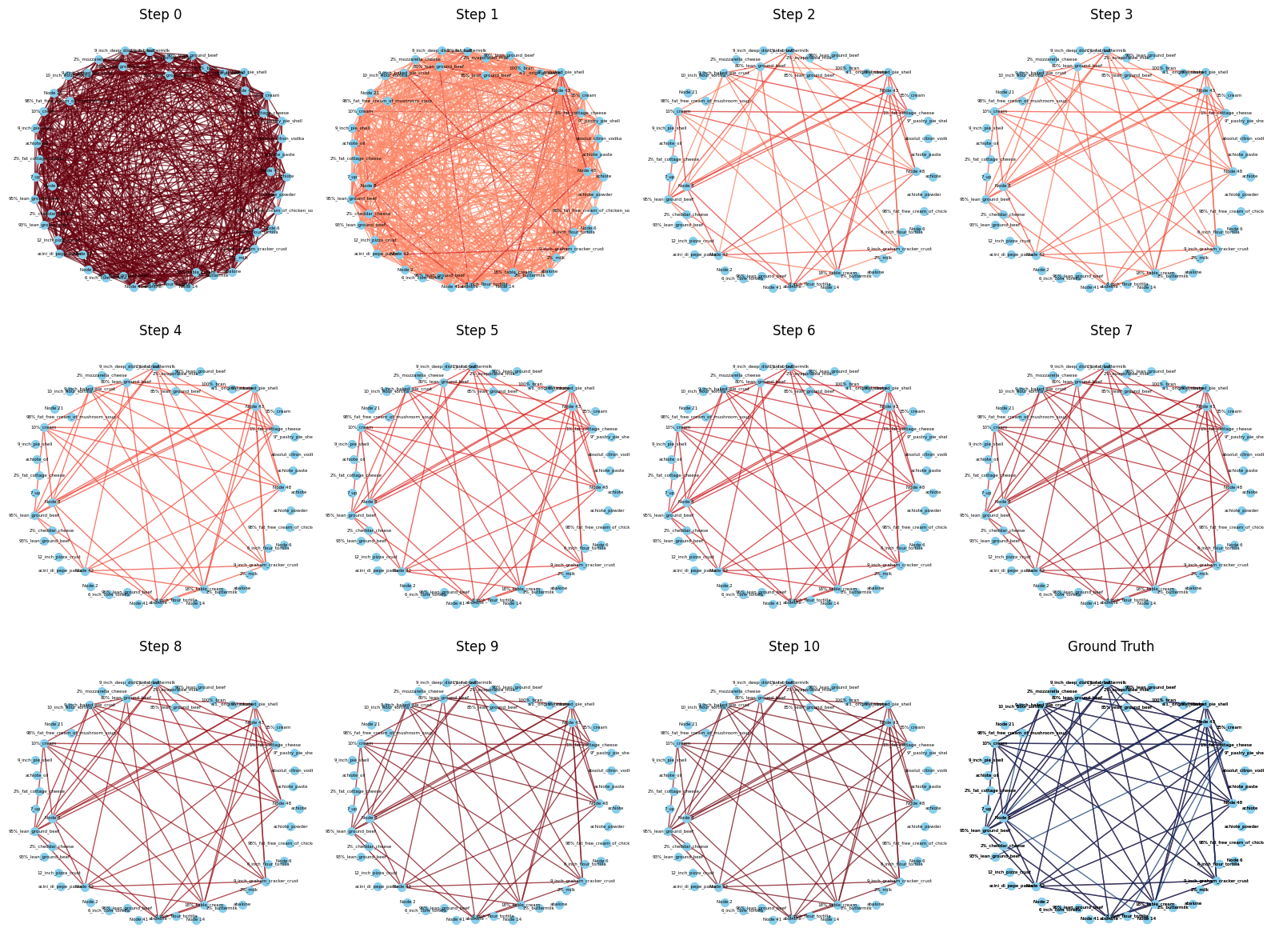}
    \caption{Progression of edge scores over diffusion steps for a 25-node subgraph. The color intensity represents edge scores normalized between 0 and 1. The reconstructed graph increasingly aligns with the ground truth structure.}
    \label{fig:generation_example}
\end{figure}

The visualization results highlight the impact of the proposed Flavor Diffusion framework on embedding quality, particularly with the CSP (Chemical Structure Prediction) layer, as shown in Figures~\ref{fig:embedding_comparison} and~\ref{fig:generation_example}.

\paragraph{Embedding Space Analysis}
Figure~\ref{fig:embedding_comparison} illustrates the differences in embedding spaces across model configurations. The baseline embeddings (left) fail to separate chemical compounds and ingredients effectively, resulting in diffuse and isotropic clusters dominated by non-hub ingredients. 

In contrast, the embeddings generated by **Flavor Diffusion (200 nodes)** without CSP (center) show improved clustering, with chemical compounds and hub ingredients forming clearer groups. However, some overlap persists between hubs and non-hubs. The inclusion of the CSP layer (right) further refines the embeddings, creating well-structured, anisotropic clusters that reflect meaningful relationships between ingredients and compounds.

\paragraph{Performance and Structural Insights}
The CSP layer significantly enhances clustering performance, as evidenced by the highest NMI mean (\textbf{0.3410}) achieved by Flavor Diffusion (200 nodes) with CSP. The transition from isotropic to anisotropic embedding spaces reflects the model's ability to learn diverse, domain-specific relationships. Furthermore, the iterative refinement process highlights the framework's capacity to generate realistic ingredient-ingredient graphs that align with culinary and chemical properties.

\paragraph{Dynamic Reconstruction for Novel Insights}
The iterative reconstruction process visualized in Figure~\ref{fig:generation_example} showcases the Flavor Diffusion framework's ability to refine ingredient-ingredient relationships progressively. Starting from random initialization (Step 0), the edge scores evolve over diffusion steps, ultimately converging towards the ground truth structure by Step 10. The color intensity of the edges reflects their normalized scores, with higher values indicating stronger relationships. This gradual alignment with the ground truth demonstrates the model's capacity to encode meaningful relational patterns in a structured manner.

\paragraph{Potential for Ingredient Innovation}

The progressive nature of the diffusion process suggests that Flavor Diffusion is not only capable of reconstructing known ingredient-ingredient relationships but also has the potential to generalize and infer connections beyond the training data. The inclusion of chemical fingerprints and iterative edge refinement allows the model to generate plausible ingredient combinations, even in scenarios involving diverse or sparse subgraph configurations. This characteristic is particularly valuable for fields such as computational gastronomy, where discovering unique and harmonious flavor pairings is a central goal.

\paragraph{Alignment with Culinary and Chemical Properties}
The alignment of the reconstructed graphs with ground truth structures further underscores the model’s fidelity in capturing culinary and chemical properties. As the diffusion process unfolds, the model demonstrates an increasing ability to balance local (ingredient-specific) and global (chemical-based) relationships. This balance not only enhances clustering quality but also provides a robust framework for extending ingredient networks in a meaningful way.
\section{Conclusion}
This study introduced FlavorDiffusion, a diffusion model-based framework for predicting ingredient pairings and chemical interactions. The model’s capability to integrate chemical fingerprints and optimize graph embeddings resulted in improved clustering quality and predictive accuracy. Experiments revealed that the inclusion of the CSP layer significantly enhanced the representation of food-chemical relationships, achieving the highest NMI scores across various configurations. The progressive nature of the diffusion process further demonstrated the model’s ability to generalize, enabling the inference of novel ingredient combinations. By aligning culinary and chemical properties, FlavorDiffusion offers a robust tool for advancing food pairing discovery, with applications in flavor design and computational gastronomy. Future work will aim to expand dataset coverage, integrate multi-modal data, and explore new graph-sampling techniques to further enrich food science research.

\bibliographystyle{plainnat}

\end{document}